\documentclass[10pt,twocolumn,letterpaper]{article}
\pdfoutput=1
\usepackage{cvpr}              

\usepackage{algorithm}
\usepackage{algorithmic}
\usepackage{amsmath}
\usepackage{amssymb}
\usepackage{multicol}
\usepackage{multirow}
\usepackage{tabularx}
\usepackage{svg}
\usepackage{xcolor}

\newcommand{\TABVSPACE}{1}
\DeclareMathAlphabet\mathbfcal{OMS}{cmsy}{b}{n}

\newcommand{\tabNA}{---}

\newcommand{\methodname}{{DetVPCC}}

\newcommand{\pointcloud}{{P}}
\newcommand{\listofpointcloud}{{\mathbf{\mathcal{P}}}}
\newcommand{\listofbboxs}{{\mathcal{B}}}
\newcommand{\kdtree}{{\mathcal{T}}}
\newcommand{\convexhull}{{H}}
\newcommand{\idxbbox}{j}
\newcommand{\numbbox}{M}
\newcommand{\idxpoints}{i}
\newcommand{\numpoints}{N}

\newcommand{\backqp}{{q_{b}}}
\newcommand{\backqpset}{{Q_{b}}}
\newcommand{\roiqp}{{q_{r}}}

\newcommand{\roimaskpc}{{\mathbfcal{R}}}
\newcommand{\roimaskimg}{\mathbf{R}}
\newcommand{\accfunc}{{\mathcal{F}}}

\newcommand{\framel}{{L}}
\newcommand{\framepixnum}{{M}}
\newcommand{\pointnum}{N}
\newcommand{\framembnum}{B}

\newcommand{\ptpixmap}{{\mathbf{M^p}}}
\newcommand{\pixmbmap}{{\mathbf{M^b}}}

\newcommand{\heatmap}{{Y}}
\newcommand{\pillarhight}{{h}}
\newcommand{\pillarwidth}{{w}}

\newcommand{\areabetwcurve}{\mathcal{A}}
\newcommand{\metric}{M}

\definecolor{cvprblue}{rgb}{0.21,0.49,0.74}
\usepackage[pagebackref,breaklinks,colorlinks,allcolors=cvprblue]{hyperref}

\title{DetVPCC: RoI-based Point Cloud Sequence Compression \\ for 3D Object Detection}

\author{Mingxuan Yan\thanks{Equal contribution} \qquad Ruijie Zhang\footnotemark[1] \qquad Xuedou Xiao \qquad Wei Wang\thanks{Corresponding Author} \\
Huazhong University of Science and Technology\\
}

\begin{document}
\maketitle
\begin{abstract}
While MPEG-standardized video-based point cloud compression (VPCC) achieves high compression efficiency for human perception, it struggles with a poor trade-off between bitrate savings and detection accuracy when supporting 3D object detectors. This limitation stems from VPCC's inability to prioritize regions of different importance within point clouds.
To address this issue, we propose \methodname{}, a novel method integrating region-of-interest (RoI) encoding with VPCC for efficient point cloud sequence compression while preserving the 3D object detection accuracy. 
Specifically, we augment VPCC to support RoI-based compression by assigning spatially non-uniform quality levels. Then, we introduce a lightweight RoI detector to identify crucial regions that potentially contain objects. Experiments on the nuScenes dataset demonstrate that our approach significantly improves the detection accuracy. 
The code and demo video are available in supplementary materials.
\vspace{-0.4cm}
\end{abstract}    
\section{Introduction}
\label{sec:intro}

Point cloud-based 3D object detection is revolutionizing a wide range of applications, significantly enhancing capabilities in fields such as autonomous driving~\cite{fernandesPointcloudBased3D2021a, mao3DObjectDetection2023a, zengRT3DRealTime3D2018} and 3D scene perception~\cite{wisultschew3DLIDARBasedObject2021, benedek3DPeopleSurveillance2014, 3DSurveillance, blanch2024lidar}. Due to the conflict between computationally intensive deep learning-based 3D object detectors and the limited computational resources of edge devices, it is promising to stream or archive point clouds to compute-capable platforms for further analytics~\cite{lianides20223d, mclean2022towards} or model optimization~\cite{yang2023online, shaheen2022continual}. However, the large volumes of point cloud data generated by sensors pose significant challenges for networking and storage.


\begin{figure}[ht]
\setlength{\abovecaptionskip}{0.2cm}
\setlength{\belowcaptionskip}{-0.2cm}
  \centering
    \includegraphics[width=\linewidth]{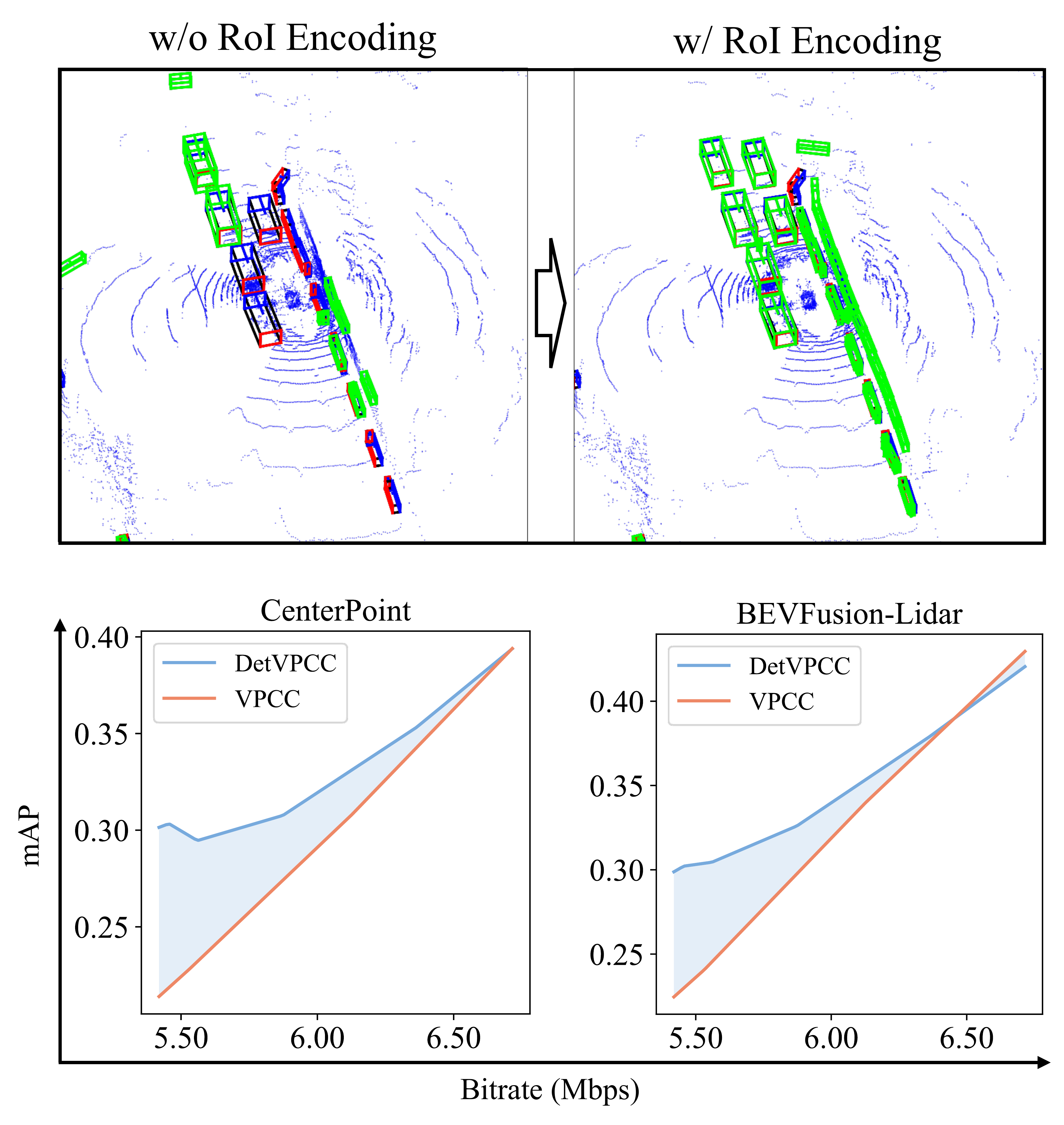}
  \caption{The first row demonstrates the accuracy improvement of 3D object detection after applying RoI encoding (objects detected are colored in green). The second row shows the accuracy-bitrate trade-off curve of \methodname{} and VPCC when supporting CenterPoint~\cite{yin2021center} and BEVFusion-Lidar~\cite{liu2023bevfusion}.\protect\footnote{}}
  \label{fig-intro-showcase}
\vspace{-0.4cm}
\end{figure}
\footnotetext{Results in the first row are encoded with nuScenes~\cite{caesarnuScenesMultimodalDataset2020} dataset, RoI QP $\roiqp=20$, background QP $\backqp=45$, CenterPoint~\cite{yin2021center} as the back-end 3D object detector. Settings of results in the second row follow Fig.~\ref{fig-metric-illus}. Details about RoI encoding can be found in ~\cref{sec-roi-encoder}.}

\begin{figure*}[t]
\setlength{\abovecaptionskip}{0.2cm}
\setlength{\belowcaptionskip}{-0.2cm}
  \centering
    \includegraphics[width=\linewidth]{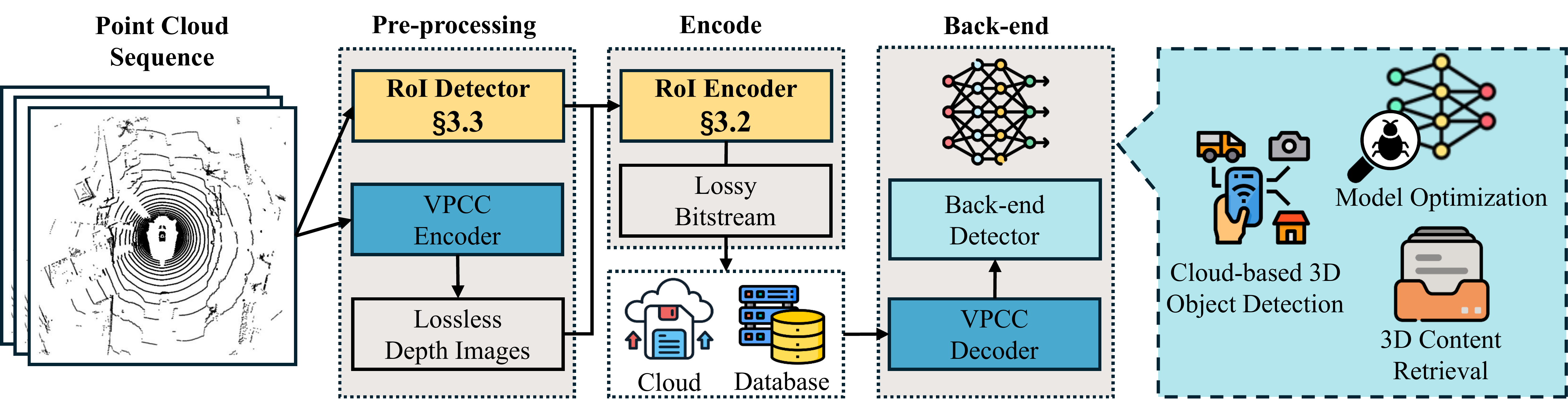}
  \caption{Overview of \methodname{}.}
  \label{fig-system-diagram}
\end{figure*}

The moving picture experts group's (MPEG) video-based point cloud compression (VPCC)~\cite{graziosi2021video} standard is a promising solution to point cloud sequence compression. VPCC projects a sequence of 3D point clouds into 2D depth and color images, and leverages mature video codecs such as H.264~\cite{wiegand2003overview}, which is widely supported by existing hardware, to remove temporal redundancy and compress the volume~\cite{graziosiOverviewOngoingPoint2020}. Typically, VPCC does not prioritize different spatial regions and applies a constant compression quality within a point cloud, which is aligned with visual quality metrics such as Point-to-Point (P2P) PSNR~\cite{mekuriaPerformanceAssessmentPoint2017} and Point-to-Plane (P2C) PSNR~\cite{tianGeometricDistortionMetrics2017a}.

However, this quality assignment scheme poses significant limitations in applications where some spatial regions, i.e., regions of interest (RoIs), demand higher fidelity. 
For instance, in driving scenes, regions containing the potential objects are crucial for 3D object detectors~\cite{qi2018frustum,paigwar2021frustum}, while the ground plane is less interested~\cite{himmelsbachFastSegmentation3D2010, leePatchworkFastRobust2022}. The uniform application of compression may result in the loss of critical details in RoIs, potentially compromising the functionality of 3D object detectors. 
To better understand this dilemma, Fig.~\ref{fig-intro-showcase} demonstrates the bitrate-accuracy curve of VPCC when applying VPCC on nuScenes dataset~\cite{caesarnuScenesMultimodalDataset2020}. As shown, VPCC experiences a poor trade-off between bitrate reduction and 3D object detection accuracy.

To this end, we propose \methodname{}, a RoI-based point cloud sequence compression method. Specifically, we enhance VPCC to support RoI encoding and design a lightweight detector to locate RoIs efficiently. As shown in Fig.~\ref{fig-intro-showcase}, with the aid of RoI encoding, our method significantly improves the accuracy of 3D object detectors. 



The contributions of this work are as follows:

\begin{itemize}
  \item To our knowledge, this is the first study that highlights and addresses the poor bitrate-accuracy trade-off of VPCC when it supports 3D object detection.
  \item We enhance VPCC to support RoI-based encoding to allow spatially non-uniform quality assignment. 
  \item We design an efficient RoI detector to locate critical regions potentially containing objects. 
  \item We evaluate \methodname{} on the nuScenes dataset. Experimental results show that \methodname{} achieves a better bitrate-accuracy trade-off than vanilla VPCC.
\end{itemize}


\section{Related Work}
\subsection{Point Cloud Sequence Compression}
Point cloud sequence compression methods can be broadly categorized into point-based and projection-based methods.
Point-based methods directly operate on 3D point clouds to remove redundant content across frames. While early methods represent points as octree~\cite{garcia2019geometry} or voxels~\cite{de2017motion}, recent approaches~\cite{gomes2021graph, akhtar2024inter} leverage deep neural networks (DNNs) to transform unstructured point clouds into latent spaces, improving coding efficiency by better correlating similar components between frames. In contrast, projection-based methods project the 3D point cloud to 2D depth images and compress the volume of depth images~\cite{graziosiOverviewOngoingPoint2020, sun2020advanced}. As part of ongoing standardization efforts, VPCC \cite{graziosiOverviewOngoingPoint2020} utilizes mature 2D video encoders like H.264 to encode depth and color images. This strategy makes VPCC one of the most promising point cloud sequence compression standards due to its seamless integration with existing 2D video encoders and hardware infrastructure.

\subsection{Encoders for Machine Vision}
As vision models consume increasing multimedia content, there is growing interest in developing machine vision-oriented encoders that aim to save bits by exploiting RoIs of vision models. For 2D video encoding, feedback-based methods~\cite{xie2019source, du2020server, li2021task, xiao2022dnn, liu2022adamask} exploits either gradient-based importance scores or predicted bounding boxes to build spatial importance maps. Neural encoder-based methods~\cite{reich2024deep, wang2022enabling} replace traditional encoders with differential neural encoders and directly optimize them to support the target vision model. On-device analytics methods~\cite{du2022accmpeg, murad2022dao, zhang2022casva} propose to use cheap analytics models to localize RoIs or dynamically control encoding parameters. For single-frame 3D point cloud encoding, Liu et al.~\cite{liu2023pchm} propose a neural encoder optimized for both human and machine perception. However, to our knowledge, none of the existing works have yet investigated 3D point cloud sequence compression. 
\section{Methodology}
\subsection{Overview}
As shown in Fig.~\ref{fig-system-diagram}, in the first stage, the RoI detector identifies RoIs within point clouds, while VPCC converts raw point clouds into lossless 2D depth images. Subsequently, the RoI encoder applies RoI-based lossy transcoding to these lossless depth images, effectively compressing the data volume. The encoded point clouds can then be archived or streamed to the cloud for further analysis by back-end 3D object detectors. This enables various applications, such as cloud-based 3D object detection, fault detection, optimization of on-device 3D object detectors, and efficient 3D content retrieval.

\subsection{RoI-based point cloud Sequence Encoder}
\label{sec-roi-encoder}
\subsubsection{Preliminary on VPCC}
To encode a point cloud sequence, VPCC first clusters point clouds into small patches grouped by point-wise surface normals. Then, it projects and arranges them into three sequences of 2D images, namely attribute, occupancy, and geometry images~\cite{graziosiOverviewOngoingPoint2020}. 
Attribute images store color and auxiliary information. Occupancy images are binary images indicating the occupancies in geometry images.
Geometry (Depth) images encode point positions into pixel values proportional to the distance between the projection plane and correlated points, as shown in Fig.~\ref{fig-roi-encoding}.
These 2D images are then compressed to separate bitstreams using standard 2D video codecs like H.264. Typically, geometry images account for most of the volume ~\cite{rudolph2023rabbit}.

To reconstruct point clouds, the compressed bitstreams are decoded and projected back to 3D point clouds using projection information encoded in the bitstream.

\subsubsection{Bitrate Control and RoI Encoding}
VPCC provides bitrate control options at both point and image levels. For the point-level bitrate control, VPCC provides options to limit the maximum number of points in a patch and the maximum distance between a selected point to the projection plane~\cite{graziosiOverviewOngoingPoint2020}. At the image level, bitrate control follows the underlying 2D video encoder. 

Following the practice of existing works~\cite{rudolph2023rabbit, shen2021rate, wang2023vqba}, we control bitrate at the image level by \textit{transcoding the lossless geometry (depth) images} with H.264~\cite{wiegand2003overview} video encoder to take advantage of the mature bitrate control mechanisms of 2D video encoders. Unlike previous works that uniformly compress geometry images, we implement non-uniform quality assignments by leveraging the macroblock-level quality control feature of 2D video encoders.

\begin{figure}[t]
\setlength{\abovecaptionskip}{0.2cm}
\setlength{\belowcaptionskip}{-0.2cm}
  \centering
    \includegraphics[width=\linewidth]{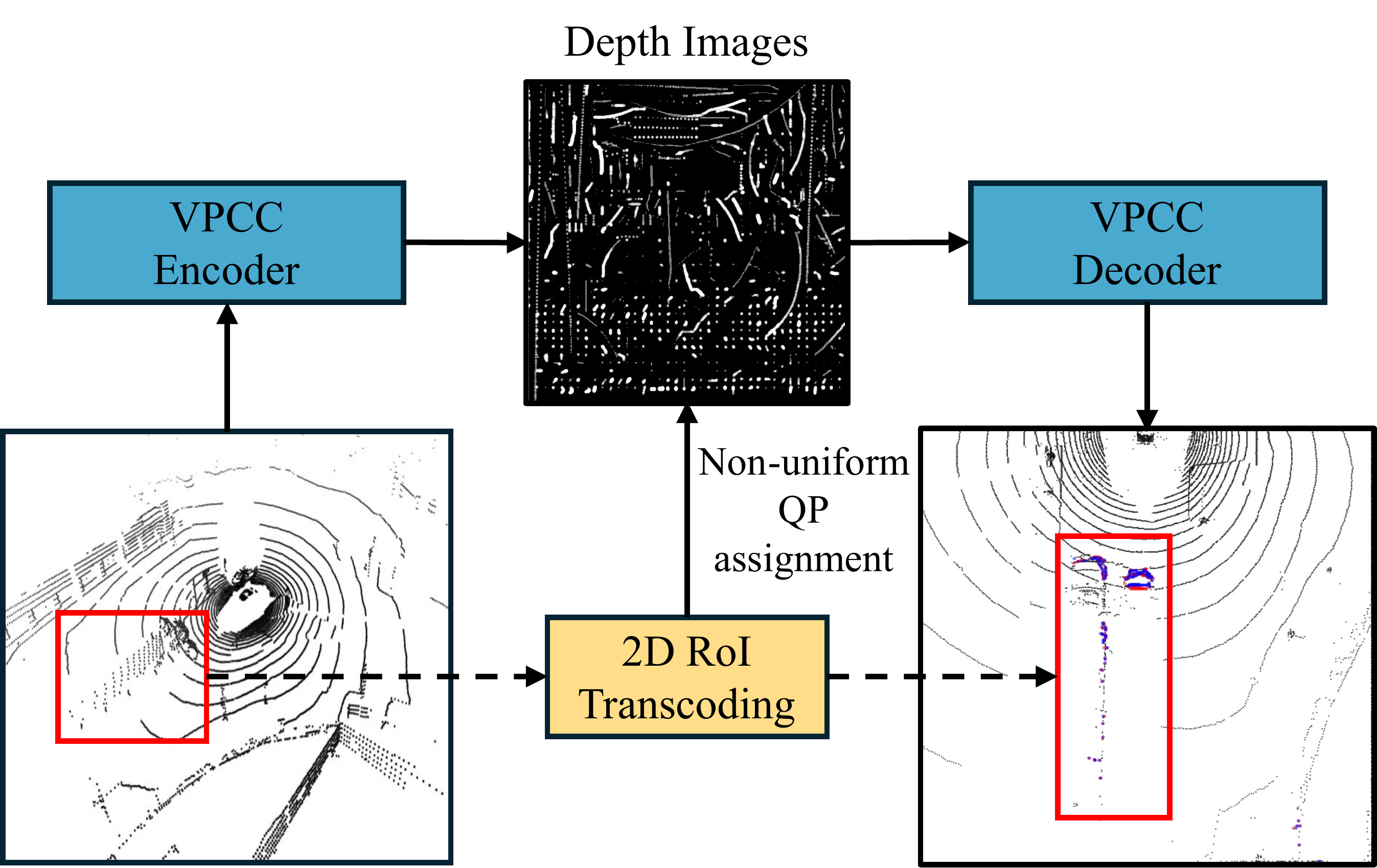}
  \caption[]{Illustration of RoI Encoding.\protect\footnote{}\ The right image compares point clouds with RoI and non-RoI encoding. The red points represent the points that appear in the non-RoI encoded frame but not in the RoI encoded frame, and blue represents the contrary. Black points represent common parts.}
  \label{fig-roi-encoding}
\end{figure}
\footnotetext{For better visualization, the depth image is cropped, and its sharpness is lowered by 25\%; The sample is encoded with $\roiqp=25$ and $\backqp=30$.}

Specifically, H.264 processes geometry images in units of macroblocks consisting of $16\times16$ pixels. Each macroblock will be further partitioned and transformed into the frequency domain by $4\times4$ 2D discrete cosine transform (DCT)~\cite{wiegand2003overview}:
\begin{align*}
    \mathbf{Y}&=\mathbf{A}\mathbf{X}\mathbf{A}^T
\end{align*}
where $\mathbf{X}$ is the $4\times4$ pixel matrix, $\mathbf{Y}$ is the DCT coefficient matrix, $\mathbf{A}$ is the orthogonal DCT transform matrix.
The above equation could be transformed into the following equivalent form~\cite{Richardson2003H2P}:

\vspace{-0.4cm}
\begin{align*}
    \mathbf{Y}=(\mathbf{C}\mathbf{X}\mathbf{C}^T)\otimes\mathbf{E}
\label{eq-dct-1}
\end{align*}
where $\mathbf{E}$ is a scaling matrix, and $\otimes$ represents element-wise multiplication. $\mathbf{W}=\mathbf{C}\mathbf{X}\mathbf{C}^T$ is the unscaled coefficient matrix, which is quantized and scaled by:
\begin{equation}
\mathbf{Z}_{ij}=round\left(\mathbf{W}_{ij}\frac{\mathbf{M}_{ij}}{2^{(15+floor(\frac{QP}{6})}}\right)
\label{eq-dct-1}
\end{equation}
where $\mathbf{Z}$ is the quantized frequency coefficient matrix, $\mathbf{M}$ is a scaling matrix derived from $\mathbf{E}$. QP, i.e., the quantization parameter, is an integral value that controls bitrate by adjusting the quantization step. \textit{The larger the QP is, the less frequency information is reserved during the quantization, the lower the image quality, and the smaller the volume.} In H.264, QP ranges from 0 to 51.

To implement non-uniform QP assignments, we use the emphasis map feature provided by NVENCODE API~\cite{NVENCVideoEncoder}. This feature allows \textit{macroblock-level control of QP}, which enables the encoder to enhance the quality of geometry images in specified partitions. As illustrated in Fig.~\ref{fig-roi-encoding}, by setting a lower QP for RoIs of depth images, the quality of RoIs in the decoded point clouds is enhanced. In \methodname{}, we use a binary quality assignment strategy by applying two distinct QPs—a low QP for RoIs and a high QP for non-RoI (background) areas.

\begin{figure*}[ht]
\setlength{\abovecaptionskip}{0.2cm}
\setlength{\belowcaptionskip}{-0.2cm}
  \centering
    \includegraphics[width=0.9\linewidth]{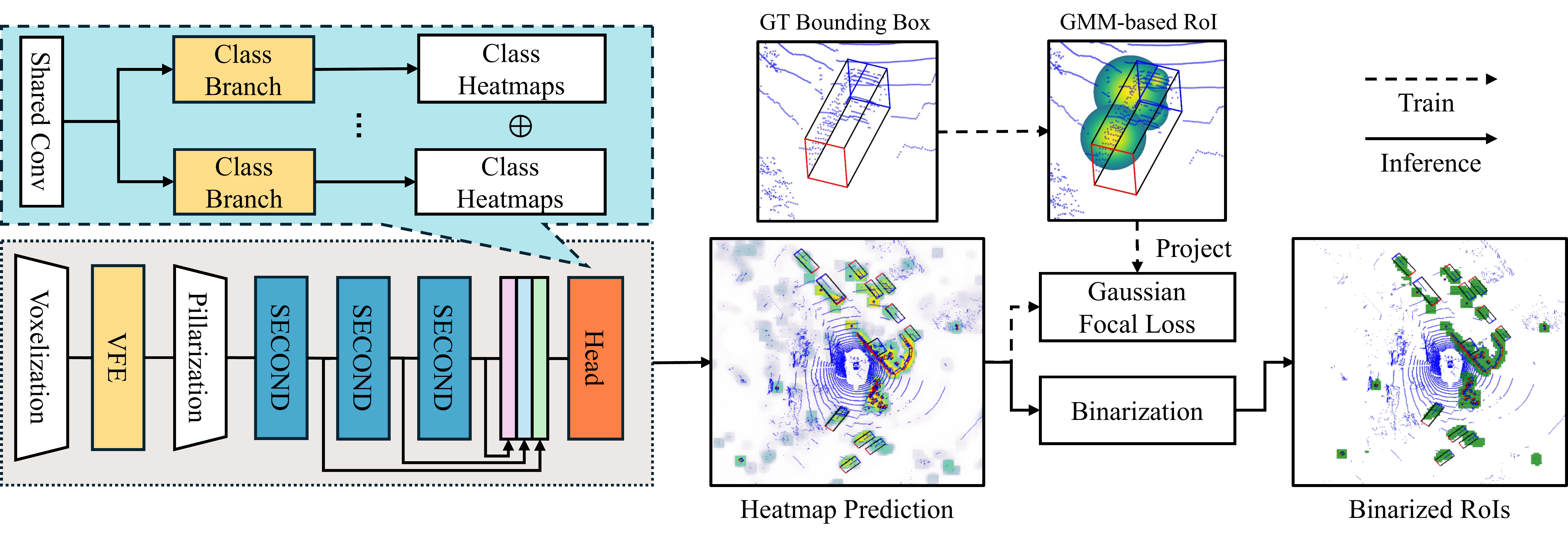}
  \caption{Design of the GMM-based RoI detector.}
  \label{fig-roi-detector}
\end{figure*}

\subsubsection{Encoding Objective}
\label{sec-encoding-objective}
The objective of the RoI encoder can be formalized as:
\begin{equation}
\max_{\roimaskimg}\sum_{\backqp\in\backqpset}\accfunc(\backqp;\roiqp,\roimaskimg)
\label{eq_encobj}
\end{equation}
$\roimaskimg\in \{0,1\}^{\framembnum \times \framel}$ is the binary RoI macroblock indicator where $\framembnum, \framel$ denote the number of macroblocks in each frame and total number of frames of the geometry image sequence to be encoded; $\accfunc$ denotes the 3D object detector's accuracy function;  $\roiqp$ is the pre-defined QP of RoI; the non-RoI (background) QP $\backqp$ is sampled from a pre-defined QP value set $\backqpset$ to comprehensively evaluate $\roimaskimg$ under different background point cloud qualities.

Typically, RoIs are detected in 3D space. A RoI mask of a given point cloud can be denoted by $\roimaskpc \in \{0,1\}^{\pointnum}$ where $\pointnum$ is the number of points in the point cloud and $\roimaskpc[n]=1$ indicates the $n^{th}$ point belongs to RoI.  Typically, $\pointnum>\framembnum$. $\roimaskimg$ can be solved by:
\begin{equation}
    \ptpixmap_t\pixmbmap_t\roimaskimg[:,t] = {\roimaskpc_t}^T 
\end{equation}
where $\framepixnum$ is the number of pixels in the frame; $t$ is the frame index; $\ptpixmap \in \{0,1\}^{\pointnum \times \framepixnum}$ is the VPCC's point-to-pixel map encoded within the bitstream and $\pixmbmap \in \{0,1\}^{\framepixnum \times \framembnum}$ is the pixel-to-macroblock index map.

As such, the core of solving Eq.~\ref{eq_encobj} is to identify the RoI mask $\roimaskpc$ for any given point cloud. In the next section, we will elaborate on the design of our RoI detector, which efficiently detects 3D RoIs.

\subsection{RoI Detector}
An intuitive design for the RoI detector is based on a lightweight 3D object detector, which uses the predicted 3D bounding box as the RoIs.
However, while object detectors try to precisely regress the bounding box parameters, our goal is to gain point-level importance scores \textit{without the need to distinguish between instances or predict the object classes.} This observation enables the RoI detector to be designed in a lightweight yet efficient way. 


\subsubsection{GMM-based RoI}
Inspired by previous heatmap-based object predictors~\cite{yin2021center,zhao2023ada3d,law2018cornernet,zhou2019objects}, the RoI detector is designed to predict a heatmap indicating the possibility an object appears in a given region. 

As shown in Fig.~\ref{fig-roi-detector}, due to irregular object shapes and occlusion, points belonging to an object tend to be \textit{distributed in clusters}, and each cluster could be interpreted as a distinct component of that object. Thus, instead of using simple Gaussian distributions to model the object centers~\cite{yin2021center, zhou2019objects}, we propose to model the points of objects with 3D Gaussian mixture models (GMMs). Specifically, we assume the point cloud $\pointcloud = \{\mathbf{p}_\idxpoints\}_{\idxpoints=1}^\numpoints$ of an object is sampled from a 3D GMM:
\begin{equation}
\mathbf{p}_\idxpoints \sim \sum_{c=1}^{C} \pi_c \, \mathcal{N}(\mathbf{p} \mid \boldsymbol{\mu}_c, \boldsymbol{\Sigma}_c)
\end{equation}
where $C$ is the number of Gaussian components in the mixture. $\pi_c$ is the weight of the $c$-th Gaussian component, with $\sum_{c=1}^C \pi_k = 1$ and $\pi_c \geq 0$. We let $\pi_c=\frac{1}{C},\forall c\in\{1,\cdots,C\}$ as a general assumption for the priors. At training time, the component centers $\boldsymbol{\mu}_c \in \mathbb{R}^3$ and covariances $\boldsymbol{\Sigma}_c \in \mathbb{R}^{3 \times 3}$ is fitted to the points located inside of each ground-truth bounding box. 

\begin{algorithm}[t]
\caption{Find Points in Bounding Boxes}
\begin{algorithmic}[1]
\REQUIRE Point cloud $\pointcloud = \{\mathbf{p}_\idxpoints\}_{\idxpoints=1}^\numpoints$; A set of bounding boxes $\listofbboxs = \{B_\idxbbox\}_{\idxbbox=1}^\numbbox$, where $B_\idxbbox$ is defined by width $w_\idxbbox$, length $l_\idxbbox$, height $h_\idxbbox$ and box center $\mathbf{c}_\idxbbox$.
\ENSURE List of inner point clouds $\listofpointcloud_{in}=\{\pointcloud_\idxbbox\}_{\idxbbox=1}^\numbbox$
\STATE $\listofpointcloud_{in} \leftarrow \emptyset$
\STATE $\kdtree \leftarrow BuildKDTree(\pointcloud)$
\FOR{each bounding box $B_\idxbbox \in \listofbboxs$}
    \STATE $r_\idxbbox=\frac{1}{2}\sqrt{w_\idxbbox^2 + l_\idxbbox^2 + h_\idxbbox^2 }$
    \STATE $\pointcloud_{r} \leftarrow \kdtree.QueryBallPoints(\mathbf{c}_\idxbbox,r_\idxbbox)$
    \STATE $\convexhull_\idxbbox \leftarrow Delaunay(Corners(B_\idxbbox))$
    \STATE $\pointcloud_\idxbbox \leftarrow \convexhull_\idxbbox.PointsInAnySimplex(\pointcloud_r)$
    \STATE $\listofpointcloud_{in} \gets \listofpointcloud_{in} \cup \{\pointcloud_\idxbbox\}$
\ENDFOR
\RETURN $\listofpointcloud_{in}$
\end{algorithmic}
\label{alg-points-in-roi}
\end{algorithm}

To accelerate the labeling process, we provide an efficient algorithm to find points inside a set of 3D bounding boxes in Alg.~\ref{alg-points-in-roi}. The algorithm essentially leverages a reusable K-D tree to narrow the search space of each bounding box down to its circumscribed sphere. Then, we adopt Quickhull algorithm~\cite{barber1996quickhull} to accelerate the searching process of finding points inside the sphere. Specifically, it performs Delaunay triangulation to separate the bounding box into a set of simplexes $\convexhull$, i.e., tetrahedrons in 3D space. Then, Alg.~\ref{alg-points-in-roi} iterate through $\pointcloud_{r}$ and check whether the point falls in any of the simplexes and get the points inside of the RoI bounding boxes $\listofpointcloud_{in}$.

To enable a lightweight network design of the RoI detector, we transform 3D GMMs to 2D heatmaps $Y\in \mathbb{R}^{\pillarhight\times\pillarwidth}$. Specifically, we first project 3D GMMs on the x-y plane. Then, for heatmap grid $(i,j)$, its value is set to the maximum sampled probability across GMMs:
\begin{equation}
    \heatmap_{ij} = \max_{k} \sum_{c=1}^{C}  \frac{1}{C} \,  \mathcal{N} (\boldsymbol{\mu}_{ij} \mid \boldsymbol{\mu}^{'}_{kc}, \boldsymbol{\Sigma}^{'}_{kc})
\end{equation}
where $k\in\{1,\cdots,K\}$, $K$ is the total number of GMMs in the frame. $\boldsymbol{\mu}_{ij}$ is the center coordinate of the grid $(i,j)$, $\boldsymbol{\mu}^{'}$ and $\boldsymbol{\Sigma}^{'}$ are parameters of projected 2D GMMs.

\subsubsection{Network Design}

\begin{table}[t]
    \centering
    \renewcommand{\arraystretch}{\TABVSPACE} 
    \setlength{\tabcolsep}{2pt} 
    \caption{Complexities of RoI detectors.}
    \begin{tabular}{c|c|c|c|c}
    \hline
     Model      & \multicolumn{2}{c|}{GMM-based RoI} & \multicolumn{2}{c}{Naive RoI} \\
     
    \hline
    Method  & FLOPs (G) & Param (M) &  FLOPs (G) & Param (M)  \\ 
\cline{2-3}         \cline{4-5}
    \hline
        Backbone & 28.51 & 4.21 & 37.50 & 4.26 \\ 
        Header & 20.49 & 0.51 & 28.22 & 1.72 \\ 
        Total & 49.00 & 4.72 & 65.72 & 5.98 \\ 
    \hline
    \end{tabular}
    \label{tab-detector-capacity}
    \vspace{-0.4cm}
\end{table}

As shown in Fig.~\ref{fig-roi-detector}, we borrowed the backbone of PointPillar~\cite{lang2019pointpillars} due to its lightweight design and efficiency. Following the design of CenterPoint~\cite{yin2021center}, the head regresses on a pillar-wise RoI heatmap of $C\times200\times200$, where $C$ is the number of object classes. The regressed heatmap is supervised by Gaussian focal loss~\cite{law2018cornernet} separately on each class channel. 
At runtime, the heatmap $\heatmap$ produced by the network is binarized by a pre-defined threshold $\gamma$ to rule out low-confident regions. Then, the heatmap takes union across class channels to aggregate class-wise RoIs. The aggregated binary heatmap $\heatmap_b\in \{0,1\}^{200\times200}$ is then transformed to point-wise RoI mask $\roimaskpc_\heatmap$. 

Table.~\ref{tab-detector-capacity} lists the FLOPs and number of parameters of the proposed network. \textit{The overhead is below 50G FLOPs, and the number of parameters is less than 5M.} 

To suppress ground points included in $\roimaskpc_\heatmap$, we utilize the ground removal algorithm Patchwork++~\cite{lee2022patchworkpp} to generate the foreground mask $\roimaskpc_{f}$. Then, the final RoI mask is given by:
\begin{equation}
    \roimaskpc=\roimaskpc_\heatmap \cdot \roimaskpc_{f}
\end{equation}

\begin{table*}[t]
    \centering
    \renewcommand{\arraystretch}{\TABVSPACE} 
    \setlength{\tabcolsep}{3pt} 
    \caption{Comparison of RoI-based methods over vanilla VPCC. Outperforming results are highlighted in bold.}
    \begin{tabular}{c|c|c|c|c|c|c|c|c|c|c|c|c}
    \hline
     Detector      & \multicolumn{6}{c|}{CenterPoint} & \multicolumn{6}{c}{BEVFusion-Lidar} \\
     
    \hline
    \multirow{2}{*}{Method}  & \multicolumn{3}{c|}{Naive RoI} & \multicolumn{3}{c|}{\methodname{}} &  \multicolumn{3}{c|}{Naive RoI} & \multicolumn{3}{c}{\methodname{}} \\ 
                               \cline{2-6}         \cline{5-7}            \cline{8-10}          \cline{11-13}
                &$\roiqp=15$&$\roiqp=20$&$\roiqp=25$ &$\roiqp=15$&$\roiqp=20$&$\roiqp=25$ &$\roiqp=15$&$\roiqp=20$&$\roiqp=25$ &$\roiqp=15$&$\roiqp=20$&$\roiqp=25$\\
    \hline
        $\areabetwcurve_{mAP}\uparrow$ & -0.26 & 1.34 & 1.94 & \textbf{1.92} & \textbf{3.51} & \textbf{3.61} & -1.61 & 0.21 & 1.24 & \textbf{0.29} & \textbf{2.33} & \textbf{2.80} \\ 
        $\areabetwcurve_{NDS}\uparrow$ & -1.00 & 1.20 & 2.26 & \textbf{1.34} & \textbf{2.90} & \textbf{3.69} & -1.44 & -0.18 & 0.75 & \textbf{0.33} & \textbf{1.65} & \textbf{2.23} \\ 
        $\areabetwcurve_{ATE}\downarrow$ & -0.22 & -2.81 & -3.80 & \textbf{-1.75} & \textbf{-3.92} & \textbf{-4.40} & \textbf{0.49} & -0.05 & -0.38 & 0.66 & \textbf{-0.07} & \textbf{-0.84} \\ 
        $\areabetwcurve_{ASE}\downarrow$ & -0.72 & -1.92 & -2.82 & \textbf{-1.44} & \textbf{-2.50} & \textbf{-3.32} & \textbf{-0.02} & \textbf{-0.37} & \textbf{-0.41} & 0.09 & 0.06 & -0.24 \\ 
        $\areabetwcurve_{AOE}\downarrow$ & 4.56 & 1.64 & \textbf{-3.54} & \textbf{1.35} & \textbf{-0.27} & -0.95 & 0.00 & -0.73 & 0.28 & \textbf{-2.14} & \textbf{-1.21} & \textbf{-0.26} \\ 
        $\areabetwcurve_{AVE}\downarrow$ & 5.60 & 0.60 & 0.56 & \textbf{0.30} & \textbf{-2.13} & \textbf{-5.89} & 5.63 & 4.11 & -0.25 & \textbf{-0.01} & \textbf{-2.73} & \textbf{-5.58} \\ 
        $\areabetwcurve_{AAE}\downarrow$ & -0.56 & \textbf{-2.81} & -3.32 & \textbf{-2.27} & -2.65 & \textbf{-4.29} & 0.27 & -0.06 & -0.57 & \textbf{-0.44} & \textbf{-0.90} & \textbf{-1.34} \\ 
    \hline
    \end{tabular}
    \label{tab-main-results}
    \vspace{-0.2cm}
\end{table*}

\section{Experiments}
\subsection{Experiment Settings}

\subsubsection{Dataset}
\label{sec-dataset}
Experiments are conducted on the nuScenes~\cite{caesarnuScenesMultimodalDataset2020} dataset, which comprises 1000 autonomous driving scenes, each lasting 20 seconds, collected in urban environments such as Boston and Singapore. Each scene sample includes data from a lidar sensor, which provides high-resolution 3D point clouds. The point clouds are sampled by a frequency of 20~Hz and annotated by a frequency of 2~Hz with 3D bounding boxes for 10 classes of objects. 

For 3D object detection, nuScenes assigns 750 scenes for training, 150 scenes for validation, and 100 for testing. However, the annotations for the test set nuScenes dataset are not directly accessible~\cite{nusceneswebsite}. As we need to conduct a vast number of rounds of evaluation on the test set, we re-split the training set to 525 for training and 125 for validation, and the original 150 scenes in the validation set are used for testing. The new splitting is applied to the training of both RoI detectors and back-end detectors. 

As frames in the same scenes have close object distributions, we select clips of 40-point cloud frames indexing from 90 to 130 from each scene to accelerate the encoding process. Note that the input to the RoI detector is the enhanced point cloud that aggregates 10 historical lidar frames following common practices~\cite{mmdet3d2020}. Thus, RoI detectors predict RoI mask $\roimaskpc$ on 4 out of the 40 frames. The RoI mask for the rest of the frames is transformed from $\roimaskpc$ based on the ego-motion provided by nuScenes.

\subsubsection{Back-end Detectors}
\label{back-end-detectors}
We select the leading 3D object detectors, CenterPoint~\cite{yin2021center} and BEVFusion-Lidar~\cite{liu2023bevfusion} as our back-end detectors. We first pre-train the detectors using the dataset splitting in ~\cref{sec-dataset}. To accommodate the back-end detectors with the compressed dataset, we fine-tune the pre-trained models on a mixed dataset comprising both lossy and lossless point clouds for one epoch. Details can be found in Suppl.~\ref{sup-backend-detector}.

\subsubsection{Baselines}
We provide two baselines for the evaluation:

\textbf{Vanilla VPCC}: The original VPCC using a uniform compression scheme. The bitrate control is also implemented during the transcoding phase, except the whole point cloud is encoded with non-RoI QP $\backqp$.

\textbf{Naive RoI}: Replace the GMM-based RoI detector of \methodname{} with a naive RoI detector directly built based on the CenterPoint~\cite{yin2021center} 3D object detector. The output 3D bounding boxes are transformed into point-level RoI indexes at the test time. The detector is trained on the same dataset described in ~\cref{sec-dataset}. Ground removal and RoI encoding follow the same procedure as \methodname{}.

To fairly compare the GMM-based RoI detector and naive RoI detector, the network capacity of the two detectors is comparable, as shown in Table.~\ref{tab-detector-capacity}. The FLOPs and parameter number of the naive RoI detector are slightly larger than the GMM-based RoI detector.

\subsubsection{Implementation Details}
\label{sec-implementation-details}
We set the number of GMM components $K=5$ and the test-time heatmap binarization threshold $\gamma=0.4$. We use MPEG-TMC2 test model v15.0~\footnote{https://github.com/MPEGGroup/mpeg-pcc-tmc2} as the base implementation of VPCC. Details about the implementation and training of the RoI detector can be found in Suppl.~\ref{sup-roi-detector}.

\subsubsection{Metrics}

\begin{figure}[t]
  \centering
    \includegraphics[width=\linewidth]{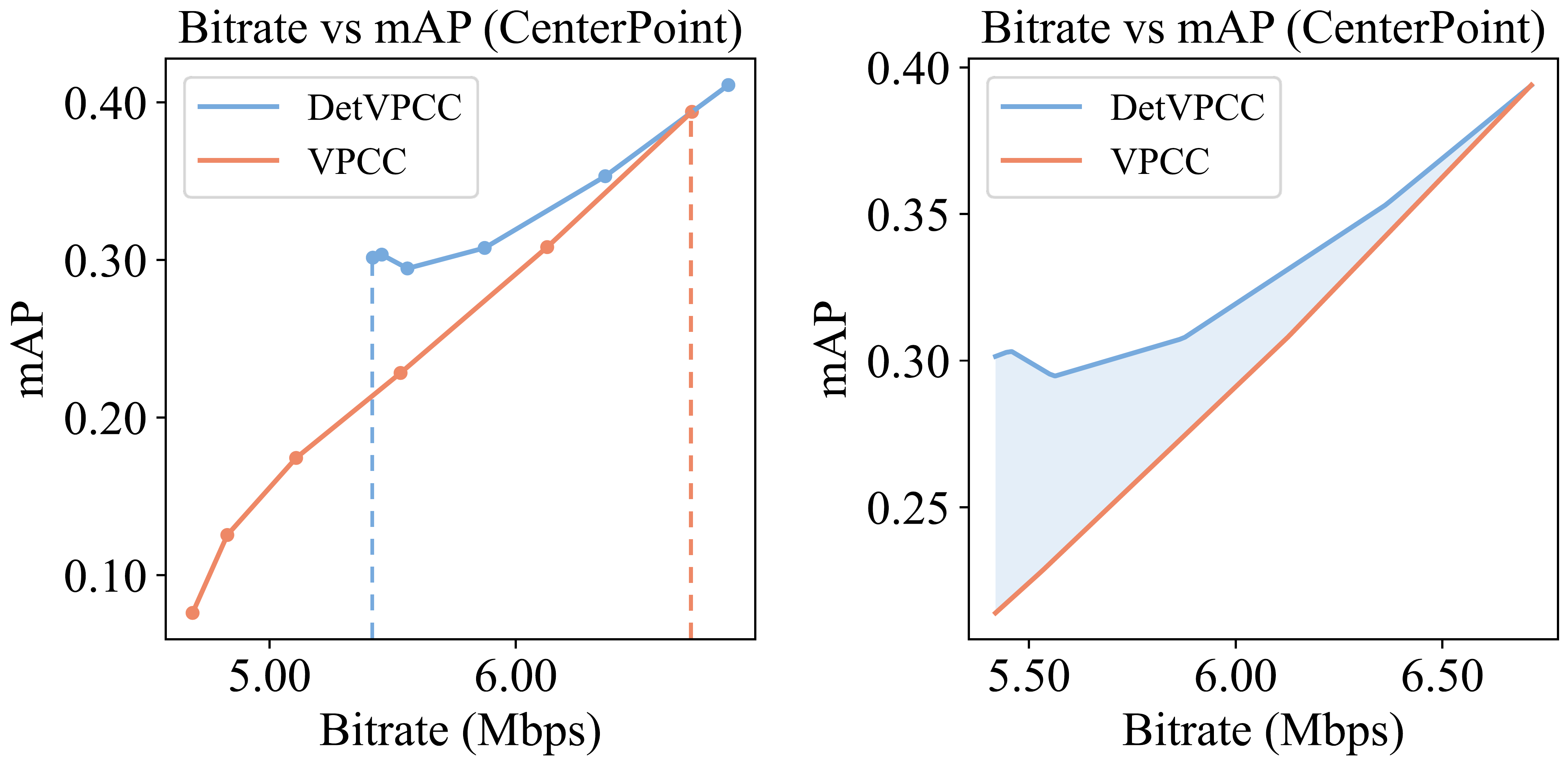}
  \caption[]{Illustration of the evaluation metric.\protect\footnote{}}
  \label{fig-metric-illus}
\vspace{-0.4cm}
\end{figure}
\footnotetext{Result is encoded with $\roiqp=25$, the rest of the settings follows \cref{sec-implementation-details}.}

For each $\roiqp\in\{15,20,25\}$, we iterate $\backqp\in\{30,33,36,39,42,45\}$ to encode the test set. Then, we run back-end 3D object detectors on the encoded test set and record their performances. The performance metrics~\cite{caesarnuScenesMultimodalDataset2020} include mean average precision (mAP), nuScenes detection score (NDS), and true-positive metrics: average translation error (ATE), average scale error (ASE), average orientation error (AOE), average velocity error (AVE) and average attribute error (AAE). We also record the averaged bitrate of encoded bitstreams across test scenes. As shown in Fig.~\ref{fig-metric-illus}, we can draw metric-bitrate curves where the y-axis is the performance metric, and the x-axis is the averaged bitrates in megabits per second (Mbps). 

We define \textit{averaged advantage} $\areabetwcurve$ to evaluate the performance boost brought by RoI encoding against vanilla VPCC:
\begin{align}
    \areabetwcurve &= \frac{\int_{x_{min}}^{x_{max}} \metric(x) - \metric_{b}(x) \, dx }{x_{max}-x_{min}}\\
                   &\approx \frac{1}{N} \sum_{i=1}^{N} \metric(x_i) - \metric_{b}(x_i) \nonumber
\end{align}
where $\metric$ and $\metric_{b}$ are the metric-bitrate curves of the RoI-based method and vanilla VPCC, respectively. $ x_{min} $ and $ x_{max} $ define the region on the x-axis where the domains of the two curves overlap. We linearly interpolate the metric-volume line chart to approximate $\areabetwcurve$, where $N=100$ is the number of samples after interpolation. In the following experiments, $\areabetwcurve$ is recorded in percentage value.

\begin{table*}[t]
\centering
\renewcommand{\arraystretch}{\TABVSPACE} 
\setlength{\tabcolsep}{6pt} 
\caption{Class-wise performance of \methodname{} over vanilla VPCC. Results outperforming VPCC are highlighted in bold.}
\begin{tabular}{c|c|c|c|c|c|c|c|c|c|c}
\hline
    Class & Car & Truck & Bus & Trailer & Constr Veh & Pedestrian & Motorcycle & Bicycle & Traffic Cone & Barrier \\ 
    \hline
    $\areabetwcurve_{mAP}\uparrow$ & -0.28 & \textbf{2.87} & \textbf{2.53} & \textbf{4.85} & \textbf{1.13} & \textbf{10.00} & \textbf{1.79} & -0.50 & \textbf{1.05} & \textbf{5.76} \\ 
    $\areabetwcurve_{ATE}\downarrow$ & \textbf{-0.09} & 0.81 & 0.25 & \textbf{-1.53} & \textbf{-1.85} & \textbf{-0.57} & \textbf{-0.51} & \textbf{-13.19} & \textbf{-1.31} & \textbf{-1.97} \\ 
    $\areabetwcurve_{ASE}\downarrow$ & \textbf{-0.13} & 0.07 & 0.40 & \textbf{-1.02} & \textbf{-1.76} & 0.26 & \textbf{-0.06} & \textbf{-9.42} & \textbf{-0.25} & \textbf{-0.30} \\ 
    $\areabetwcurve_{AOE}\downarrow$ & \textbf{-0.66} & \textbf{-1.03} & \textbf{-0.25} & 7.39 & \textbf{-3.95} & \textbf{-2.39} & 2.96 & \textbf{-8.48} & \tabNA & \textbf{-0.25} \\ 
    $\areabetwcurve_{AVE}\downarrow$ & \textbf{-0.84} & \textbf{-1.97} & \textbf{-3.94} & \textbf{-12.09} & 0.14 & \textbf{-1.32} & 2.47 & \textbf{-1.85} & \tabNA & \tabNA \\ 
    $\areabetwcurve_{AAE}\downarrow$ & \textbf{-0.43} & 0.04 & \textbf{-0.82} & \textbf{-1.06} & 0.49 & \textbf{-2.41} & 1.82 & \textbf{-11.83} & \tabNA & \tabNA \\ 
\hline
\end{tabular}
\label{tab-abla-class-wise-perf}
\vspace{-0.4cm}
\end{table*}

\subsection{Main Results}
\label{sec-main-results}
Table.~\ref{tab-main-results} shows the main results of the experiment. It can be seen that: 1) \textit{\methodname{} consistently improves VPCC across different RoI QPs and back-end detectors.} 2) \textit{The GMM-based RoI detector significantly outperforms naive RoI detector in most cases.} This shows that GMM-based RoI is more efficient for finding critical regions potentially containing objects. For the naive RoI detector, 3D object detection redundantly requires the network to classify the objects, which wastes the network's capacity. 3)  Compared with vanilla VPCC and naive RoI, \textit{\methodname{} consistently improves the true-positive metrics.} This implies that with GMM-based RoI encoding, the details of objects are well-preserved so that the back-end detector can easily identify the boundary and pose of objects. 4) \textit{The advantage over vanilla VPCC for both RoI-based methods enlarges when $\roiqp$ increases.} This is partly because when the RoI quality is degraded, the domain of the metric-bitrate curves of RoI-based methods will shift towards low bitrates, where the advantage of RoI encoding is more significant. However, when $\roiqp$ is too large, the absolute performance of the RoI-based methods will be less satisfying. Thus, we empirically keep $\roiqp$ no larger than 25.

\subsection{Additional Experiments}
We conduct additional experiments to get a detailed understanding of \methodname{}'s performance. Unless specially mentioned, we set $\roiqp=20$, $K=5$, $\gamma=0.4$, and $\areabetwcurve$ is averaged across back-end detectors. 

\vspace{-0.2cm}
\subsubsection{Class-wise Analysis}

Table.~\ref{tab-abla-class-wise-perf} demonstrates the class-wise performance of \methodname{} over vanilla VPCC. It can be seen that: 1) \textit{The smaller the object, the greater the performance improvement achieved by \methodname{} over VPCC.} For instance, the mAP advantage of \textit{pedestrain} class improves significantly by 10\%; the mAP advantages of \textit{barrier} class improves by 5.76\%; the ATE and AAE advantages of \textit{bicycle} class improve over 10\%. These results indicate that \methodname{} preserves much more object details than vanilla VPCC, given the same bitrate consumption. 2) \textit{\methodname{} help distinguish between visually similar classes.} For instance, mAP advantages of \textit{truck} and \textit{bus} both improve by approximately 3\%. 3) Interestingly, \textit{the performances of \textit{car} class are less significant compared to other classes.} It may be that the key features for identifying the \textit{car} class are more resilient to VPCC compression.

\vspace{-0.2cm}
\subsubsection{Number of GMM Components}

\begin{table}[ht]
\vspace{-0.4cm}
\centering
\renewcommand{\arraystretch}{\TABVSPACE} 
\setlength{\tabcolsep}{10pt} 
\caption{Impact of the number of GMM Components}
\begin{tabular}{c|c|c|c|c}
\hline
$K$ & 1 & 3 & 5 & 7 \\
\hline
        $\areabetwcurve_{mAP}\uparrow$ & 0.80 & 2.89 & 2.92 & 2.48 \\ 
        $\areabetwcurve_{NDS}\uparrow$ & 2.05 & 2.54 & 2.28 & 2.10 \\ 
        $\areabetwcurve_{ATE}\downarrow$ & -2.16 & -1.92 & -2.00 & -1.41 \\ 
        $\areabetwcurve_{ASE}\downarrow$ & -2.43 & -1.21 & -1.22 & -1.01 \\ 
        $\areabetwcurve_{AOE}\downarrow$ & -3.06 & -1.59 & -0.74 & -2.02 \\ 
        $\areabetwcurve_{AVE}\downarrow$ & -5.51 & -4.19 & -2.43 & -2.22 \\ 
        $\areabetwcurve_{AAE}\downarrow$ & -3.75 & -2.08 & -1.78 & -1.93 \\ 
\hline
\end{tabular}
\label{tab-abla-GMM-K}
\vspace{-0.2cm}
\end{table}

Intuitively, increasing the number of GMM components $K$ allows the RoI detector to better fit the object point clouds (the learning objective will be the same as point cloud segmentation when $K\to \infty$), so RoIs will contain fewer background points. However, as shown in Table~\ref{tab-abla-GMM-K}, the performance of \methodname{} starts to decline when $K=7$. This can be attributed to the observation that a simpler object representation tends to stabilize the learning process~\cite{yin2021center, zhou2019objects}. Additionally, the benefits of a more precise fit diminish when $K$ becomes large, as the number of background points involved in heatmaps is already minimized.

\vspace{-0.2cm}
\subsubsection{Heatmap Binarization Threshold}

\begin{table}[ht]
\vspace{-0.4cm}
\centering
\renewcommand{\arraystretch}{\TABVSPACE} 
\setlength{\tabcolsep}{12pt} 
\caption{Impact of heatmap binarization threshold}
\begin{tabular}{c|c|c|c|c}
\hline
$\gamma$ & 0.2 & 0.3 & 0.4 & 0.5 \\
\hline
        $\areabetwcurve_{mAP}\uparrow$ & -0.69 & 1.67 & 2.92 & 2.80 \\ 
        $\areabetwcurve_{NDS}\uparrow$ & -1.06 & 1.37 & 2.28 & 2.64 \\ 
        $\areabetwcurve_{ATE}\downarrow$ & 0.96 & -1.19 & -2.00 & -2.10 \\ 
        $\areabetwcurve_{ASE}\downarrow$ & 0.00 & -0.82 & -1.22 & -1.54 \\ 
        $\areabetwcurve_{AOE}\downarrow$ & 2.49 & -0.68 & -0.74 & -2.31 \\ 
        $\areabetwcurve_{AVE}\downarrow$ & 4.04 & -1.50 & -2.43 & -4.29 \\ 
        $\areabetwcurve_{AAE}\downarrow$ & -0.38 & -1.17 & -1.78 & -2.20 \\ 
\hline
\end{tabular}
\label{tab-abla-heatmap-thrd}
\vspace{-0.2cm}
\end{table}

Table.~\ref{tab-abla-heatmap-thrd} shows the impact of the heatmap binarization threshold $\gamma$. It can be seen that $\gamma=0.4$ achieves the best mAP and NDS performance. When $\gamma$ is too small, the predicted RoI regions will include too many non-RoI points, leading to poor compression efficiency. On the contrary, when $\gamma$ is too large, potential objects may be ignored, leading to low detection accuracy.
It is worth noting that $\gamma=0.5$ shows improvement on true-positive metrics. It is expected because regions with higher heatmap scores are inherently easier to identify and localize.

\begin{figure*}[t!]
\setlength{\abovecaptionskip}{0.2cm}
\setlength{\belowcaptionskip}{-0.2cm}
  \centering
    \includegraphics[width=\linewidth]{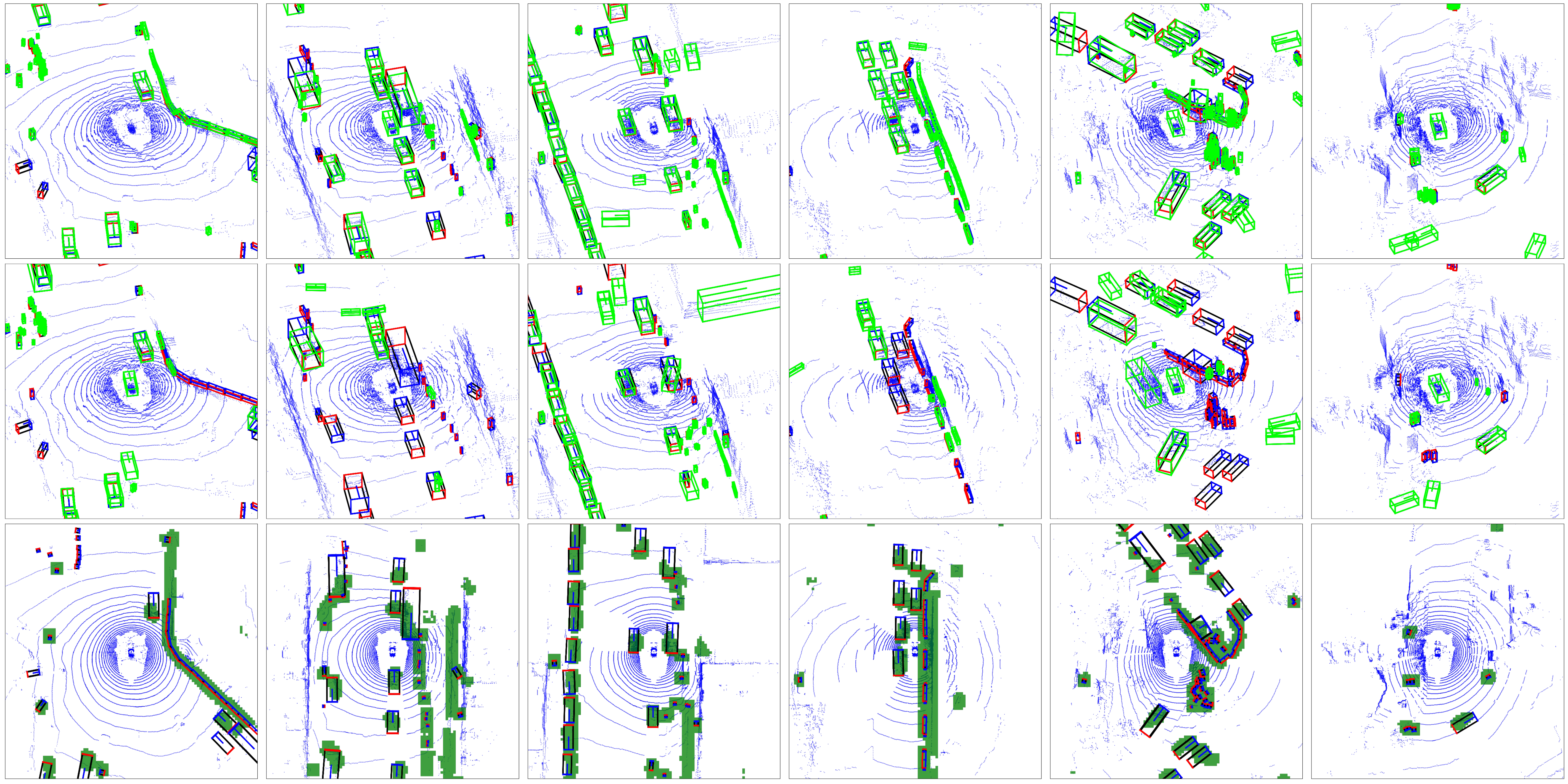}
  \caption[]{Qualitative results. The first and second rows are the detection results of CenterPoint on point clouds encoded with DetVPCC and VPCC, respectively, where $\roiqp=20$ and $\backqp=45$. The third row shows RoIs (colored in green) predicted by the RoI detector.}
  \label{fig-qualitative}
\vspace{-0.2cm}
\end{figure*}

\vspace{-0.2cm}
\subsubsection{Ablation on Ground Removal}

\begin{table}[ht]
\vspace{-0.4cm}
\centering
\renewcommand{\arraystretch}{\TABVSPACE} 
\setlength{\tabcolsep}{12pt} 
\caption{Ablation on ground removal}
\begin{tabular}{c|c|c}
\hline
Ground Removal & w/ & w/o \\
\hline
        $\areabetwcurve_{mAP}\uparrow$ & 2.92 & 1.82 \\ 
        $\areabetwcurve_{NDS}\uparrow$ & 2.28 & 1.15 \\ 
        $\areabetwcurve_{ATE}\downarrow$ & -2.00 & -0.55 \\ 
        $\areabetwcurve_{ASE}\downarrow$ & -1.22 & -0.76 \\ 
        $\areabetwcurve_{AOE}\downarrow$ & -0.74 & -0.74 \\ 
        $\areabetwcurve_{AVE}\downarrow$ & -2.43 & 0.62 \\ 
        $\areabetwcurve_{AAE}\downarrow$ & -1.78 & -0.96 \\ 
\hline
\end{tabular}
\label{tab-abla-ground-removal}
\vspace{-0.2cm}
\end{table}

Table.~\ref{tab-abla-ground-removal} shows the contribution of ground removal in \methodname{}. Ground removal assists the performance of \methodname{} to some extent as it compensates for the 2D heatmap on the z-axis and helps filter the ground points. However, DetVPCC still outperforms vanilla VPCC without ground removal, proving its effectiveness. In contrast, the naive RoI detector, which is also equipped with ground removal, exhibits compromised performance as mentioned in \cref{sec-main-results}.

\subsubsection{Find Points in Bounding Boxes}
\vspace{-0.4cm}
\begin{table}[ht]
    \centering
    \caption{Speed test of Alg.~\ref{alg-points-in-roi}}
    \begin{tabular}{c|c|c}
    \hline
        Method & Ours & nuScenes Toolkit \\ \hline
        Time per Box (ms) & 0.70 & 2.15 \\ \hline
    \end{tabular}
    \label{tab-alg1-speed}
\vspace{-0.2cm}
\end{table}

Table.~\ref{tab-alg1-speed} evaluates the speed of Alg.~\ref{alg-points-in-roi}. 
The baseline method is the \textit{points\_in\_box} function provided by nuScenes 
toolkit~\footnote{https://github.com/nutonomy/nuscenes-devkit}, 
which examines each point's relative position to the bounding box's three axes.
Experiments are conducted on Intel Xeon Gold 6226R CPUs. As shown, our method is $\sim3\times$ faster than the baseline method. The speed-up is attributed to the reusable K-D tree that drastically narrows the search space and cuts unnecessary evaluation between points and boxes that are too far away.

\subsection{Qualitative Results}

Fig.~\ref{fig-qualitative} shows visualized detection results between \methodname{} and vanilla VPCC, along with heatmaps predicted by the RoI detector. It can be seen that RoI-based encoding significantly aids 3D object detection, especially for small objects that are easily lost by vanilla VPCC.

\section{Limitations and Future Works}
While this work primarily focuses on developing an RoI-based point cloud sequence encoder for 3D object detectors, as well as an efficient RoI detector, several aspects remain for future exploration: 
1) For simplicity, pre-trained back-end 3D object detectors are fine-tuned on uniformly compressed point clouds. The unseen RoI-encoded point cloud may compromise detection accuracy. 
2) We use a constant RoI QP for all point cloud frames in a sequence. Dynamic RoI QP assignment according to the frame's complexity may further improve the compression efficiency. 
3) We assume RoIs are exactly where objects are located. However, non-object regions may also contain information critical for object identification. A more adaptive RoI design may further improve the performance.
4) Our method could be extended from autonomous driving scenes to support a broader range of applications, such as indoor 3D scene understanding~\cite{dai2017scannet, armeni_cvpr16}.

\section{Conclusion}
In this paper, we introduce \methodname{}, a RoI-based point cloud sequence compression framework. We implemented and evaluated \methodname{} on the nuScenes dataset, and the results show its superiority. We believe the proposed method can potentially alleviate the storage and network bandwidth demands of point cloud-based object detection systems through its efficient RoI-based encoding scheme.


{
    \small
    \bibliographystyle{ieeenat_fullname}
    \bibliography{main}

\begin{thebibliography}{55}
\providecommand{\natexlab}[1]{#1}
\providecommand{\url}[1]{\texttt{#1}}
\expandafter\ifx\csname urlstyle\endcsname\relax
  \providecommand{\doi}[1]{doi: #1}\else
  \providecommand{\doi}{doi: \begingroup \urlstyle{rm}\Url}\fi

\bibitem[nus()]{nusceneswebsite}
nuscenes.org.
\newblock \url{https://www.nuscenes.org/object-detection}.
\newblock [Accessed 29-10-2024].

\bibitem[Akhtar et~al.(2024)Akhtar, Li, and Van~der Auwera]{akhtar2024inter}
Anique Akhtar, Zhu Li, and Geert Van~der Auwera.
\newblock Inter-frame compression for dynamic point cloud geometry coding.
\newblock \emph{IEEE Transactions on Image Processing}, 2024.

\bibitem[Armeni et~al.(2016)Armeni, Sener, Zamir, Jiang, Brilakis, Fischer, and Savarese]{armeni_cvpr16}
Iro Armeni, Ozan Sener, Amir~R. Zamir, Helen Jiang, Ioannis Brilakis, Martin Fischer, and Silvio Savarese.
\newblock 3d semantic parsing of large-scale indoor spaces.
\newblock In \emph{Proceedings of the IEEE International Conference on Computer Vision and Pattern Recognition}, 2016.

\bibitem[Barber et~al.(1996)Barber, Dobkin, and Huhdanpaa]{barber1996quickhull}
C~Bradford Barber, David~P Dobkin, and Hannu Huhdanpaa.
\newblock The quickhull algorithm for convex hulls.
\newblock \emph{ACM Transactions on Mathematical Software (TOMS)}, 22\penalty0 (4):\penalty0 469--483, 1996.

\bibitem[Benedek(2014)]{benedek3DPeopleSurveillance2014}
Csaba Benedek.
\newblock {3D} people surveillance on range data sequences of a rotating {Lidar}.
\newblock \emph{Pattern Recognition Letters}, 50:\penalty0 149--158, 2014.

\bibitem[Blanch et~al.(2024)Blanch, Li, Escalera, and Nasrollahi]{blanch2024lidar}
Miquel~Romero Blanch, Zenjie Li, Sergio Escalera, and Kamal Nasrollahi.
\newblock Lidar-assisted 3d human detection for video surveillance.
\newblock In \emph{Proceedings of the IEEE/CVF Winter Conference on Applications of Computer Vision}, pages 123--131, 2024.

\bibitem[Caesar et~al.(2020)Caesar, Bankiti, Lang, Vora, Liong, Xu, Krishnan, Pan, Baldan, and Beijbom]{caesarnuScenesMultimodalDataset2020}
Holger Caesar, Varun Bankiti, Alex~H. Lang, Sourabh Vora, Venice~Erin Liong, Qiang Xu, Anush Krishnan, Yu Pan, Giancarlo Baldan, and Oscar Beijbom.
\newblock {nuScenes}: {A} {Multimodal} {Dataset} for {Autonomous} {Driving}.
\newblock In \emph{2020 {IEEE}/{CVF} {Conference} on {Computer} {Vision} and {Pattern} {Recognition} ({CVPR})}, pages 11618--11628, Seattle, WA, USA, 2020. IEEE.

\bibitem[Contributors(2020)]{mmdet3d2020}
MMDetection3D Contributors.
\newblock {MMDetection3D: OpenMMLab} next-generation platform for general {3D} object detection.
\newblock \url{https://github.com/open-mmlab/mmdetection3d}, 2020.

\bibitem[De~Queiroz and Chou(2017)]{de2017motion}
Ricardo~L De~Queiroz and Philip~A Chou.
\newblock Motion-compensated compression of dynamic voxelized point clouds.
\newblock \emph{IEEE Transactions on Image Processing}, 26\penalty0 (8):\penalty0 3886--3895, 2017.

\bibitem[Du et~al.(2020)Du, Pervaiz, Yuan, Chowdhery, Zhang, Hoffmann, and Jiang]{du2020server}
Kuntai Du, Ahsan Pervaiz, Xin Yuan, Aakanksha Chowdhery, Qizheng Zhang, Henry Hoffmann, and Junchen Jiang.
\newblock Server-driven video streaming for deep learning inference.
\newblock In \emph{Proceedings of the Annual conference of the ACM Special Interest Group on Data Communication on the applications, technologies, architectures, and protocols for computer communication}, pages 557--570, 2020.

\bibitem[Du et~al.(2022)Du, Zhang, Arapin, Wang, Xia, and Jiang]{du2022accmpeg}
Kuntai Du, Qizheng Zhang, Anton Arapin, Haodong Wang, Zhengxu Xia, and Junchen Jiang.
\newblock Accmpeg: Optimizing video encoding for video analytics.
\newblock \emph{arXiv preprint arXiv:2204.12534}, 2022.

\bibitem[et~al.(2017)]{dai2017scannet}
Dai et al.
\newblock Scannet: Richly-annotated 3d reconstructions of indoor scenes.
\newblock In \emph{CVPR, IEEE}, 2017.

\bibitem[Fernandes et~al.(2021)Fernandes, Silva, Névoa, Simões, Gonzalez, Guevara, Novais, Monteiro, and Melo-Pinto]{fernandesPointcloudBased3D2021a}
Duarte Fernandes, António Silva, Rafael Névoa, Cláudia Simões, Dibet Gonzalez, Miguel Guevara, Paulo Novais, João Monteiro, and Pedro Melo-Pinto.
\newblock Point-cloud based {3D} object detection and classification methods for self-driving applications: {A} survey and taxonomy.
\newblock \emph{Information Fusion}, 68:\penalty0 161--191, 2021.

\bibitem[Garcia et~al.(2019)Garcia, Fonseca, Ferreira, and De~Queiroz]{garcia2019geometry}
Diogo~C Garcia, Tiago~A Fonseca, Renan~U Ferreira, and Ricardo~L De~Queiroz.
\newblock Geometry coding for dynamic voxelized point clouds using octrees and multiple contexts.
\newblock \emph{IEEE Transactions on Image Processing}, 29:\penalty0 313--322, 2019.

\bibitem[Gomes(2021)]{gomes2021graph}
Pedro Gomes.
\newblock Graph-based network for dynamic point cloud prediction.
\newblock In \emph{Proceedings of the 12th ACM Multimedia Systems Conference}, pages 393--397, 2021.

\bibitem[Graziosi et~al.(2020)Graziosi, Nakagami, Kuma, Zaghetto, Suzuki, and Tabatabai]{graziosiOverviewOngoingPoint2020}
D. Graziosi, O. Nakagami, S. Kuma, A. Zaghetto, T. Suzuki, and A. Tabatabai.
\newblock An overview of ongoing point cloud compression standardization activities: video-based ({V}-{PCC}) and geometry-based ({G}-{PCC}).
\newblock \emph{APSIPA Transactions on Signal and Information Processing}, 9\penalty0 (1), 2020.

\bibitem[Graziosi(2021)]{graziosi2021video}
Danillo~Bracco Graziosi.
\newblock Video-based dynamic mesh coding.
\newblock In \emph{2021 IEEE International Conference on Image Processing (ICIP)}, pages 3133--3137. IEEE, 2021.

\bibitem[Hexagon(2021)]{3DSurveillance}
Hexagon.
\newblock Discover 3d surveillance, 2021.

\bibitem[Himmelsbach et~al.(2010)Himmelsbach, Hundelshausen, and Wuensche]{himmelsbachFastSegmentation3D2010}
M. Himmelsbach, Felix~v. Hundelshausen, and H.-J. Wuensche.
\newblock Fast segmentation of {3D} point clouds for ground vehicles.
\newblock In \emph{2010 {IEEE} {Intelligent} {Vehicles} {Symposium}}, pages 560--565, 2010.
\newblock ISSN: 1931-0587.

\bibitem[Kingma(2014)]{kingma2014adam}
Diederik~P Kingma.
\newblock Adam: A method for stochastic optimization.
\newblock \emph{arXiv preprint arXiv:1412.6980}, 2014.

\bibitem[Lang et~al.(2019)Lang, Vora, Caesar, Zhou, Yang, and Beijbom]{lang2019pointpillars}
Alex~H Lang, Sourabh Vora, Holger Caesar, Lubing Zhou, Jiong Yang, and Oscar Beijbom.
\newblock Pointpillars: Fast encoders for object detection from point clouds.
\newblock In \emph{Proceedings of the IEEE/CVF conference on computer vision and pattern recognition}, pages 12697--12705, 2019.

\bibitem[Law and Deng(2018)]{law2018cornernet}
Hei Law and Jia Deng.
\newblock Cornernet: Detecting objects as paired keypoints.
\newblock In \emph{Proceedings of the European conference on computer vision (ECCV)}, pages 734--750, 2018.

\bibitem[Lee et~al.(2022{\natexlab{a}})Lee, Lim, and Myung]{lee2022patchworkpp}
Seungjae Lee, Hyungtae Lim, and Hyun Myung.
\newblock {Patchwork++: Fast and robust ground segmentation solving partial under-segmentation using 3D point cloud}.
\newblock In \emph{Proc. IEEE/RSJ Int. Conf. Intell. Robots Syst.}, pages 13276--13283, 2022{\natexlab{a}}.

\bibitem[Lee et~al.(2022{\natexlab{b}})Lee, Lim, and Myung]{leePatchworkFastRobust2022}
Seungjae Lee, Hyungtae Lim, and Hyun Myung.
\newblock Patchwork++: {Fast} and {Robust} {Ground} {Segmentation} {Solving} {Partial} {Under}-{Segmentation} {Using} {3D} {Point} {Cloud}.
\newblock In \emph{2022 {IEEE}/{RSJ} {International} {Conference} on {Intelligent} {Robots} and {Systems} ({IROS})}, pages 13276--13283, 2022{\natexlab{b}}.
\newblock ISSN: 2153-0866.

\bibitem[Li et~al.(2021)Li, Shi, and Chen]{li2021task}
Xin Li, Jun Shi, and Zhibo Chen.
\newblock Task-driven semantic coding via reinforcement learning.
\newblock \emph{IEEE Transactions on Image Processing}, 30:\penalty0 6307--6320, 2021.

\bibitem[Lianides et~al.(2022)Lianides, Chan, Ismail, Harshbarger, Levorato, Callegaro, and Contreras]{lianides20223d}
Alexander Lianides, Isaac Chan, Mohamed Ismail, Ian Harshbarger, Marco Levorato, Davide Callegaro, and Sharon~LG Contreras.
\newblock 3d object detection for aerial platforms via edge computing: An experimental evaluation.
\newblock In \emph{2022 18th International Conference on Distributed Computing in Sensor Systems (DCOSS)}, pages 252--260. IEEE, 2022.

\bibitem[Liu et~al.(2023{\natexlab{a}})Liu, Hu, and Zhang]{liu2023pchm}
Lei Liu, Zhihao Hu, and Jing Zhang.
\newblock Pchm-net: A new point cloud compression framework for both human vision and machine vision.
\newblock In \emph{2023 IEEE International Conference on Multimedia and Expo (ICME)}, pages 1997--2002. IEEE, 2023{\natexlab{a}}.

\bibitem[Liu et~al.(2022)Liu, Wang, Li, Sun, Srivastava, and Abdelzaher]{liu2022adamask}
Shengzhong Liu, Tianshi Wang, Jinyang Li, Dachun Sun, Mani Srivastava, and Tarek Abdelzaher.
\newblock Adamask: Enabling machine-centric video streaming with adaptive frame masking for dnn inference offloading.
\newblock In \emph{Proceedings of the 30th ACM international conference on multimedia}, pages 3035--3044, 2022.

\bibitem[Liu et~al.(2023{\natexlab{b}})Liu, Tang, Amini, Yang, Mao, Rus, and Han]{liu2023bevfusion}
Zhijian Liu, Haotian Tang, Alexander Amini, Xinyu Yang, Huizi Mao, Daniela~L Rus, and Song Han.
\newblock Bevfusion: Multi-task multi-sensor fusion with unified bird's-eye view representation.
\newblock In \emph{2023 IEEE international conference on robotics and automation (ICRA)}, pages 2774--2781. IEEE, 2023{\natexlab{b}}.

\bibitem[Mao et~al.(2023)Mao, Shi, Wang, and Li]{mao3DObjectDetection2023a}
Jiageng Mao, Shaoshuai Shi, Xiaogang Wang, and Hongsheng Li.
\newblock {3D} {Object} {Detection} for {Autonomous} {Driving}: {A} {Comprehensive} {Survey}.
\newblock \emph{International Journal of Computer Vision}, 131\penalty0 (8):\penalty0 1909--1963, 2023.

\bibitem[McLean et~al.(2022)McLean, Xue, Lu, and Marina]{mclean2022towards}
Fraser McLean, Leyang Xue, Chris~Xiaoxuan Lu, and Mahesh Marina.
\newblock Towards edge-assisted real-time 3d segmentation of large scale lidar point clouds.
\newblock In \emph{Proceedings of the 6th International Workshop on Embedded and Mobile Deep Learning}, pages 1--6, 2022.

\bibitem[Mekuria et~al.(2017)Mekuria, Laserre, and Tulvan]{mekuriaPerformanceAssessmentPoint2017}
Rufael Mekuria, Sebastien Laserre, and Christian Tulvan.
\newblock Performance assessment of point cloud compression.
\newblock In \emph{2017 {IEEE} {Visual} {Communications} and {Image} {Processing} ({VCIP})}, pages 1--4, 2017.

\bibitem[Murad et~al.(2022)Murad, Nguyen, and Yan]{murad2022dao}
Taslim Murad, Anh Nguyen, and Zhisheng Yan.
\newblock Dao: Dynamic adaptive offloading for video analytics.
\newblock In \emph{Proceedings of the 30th ACM International Conference on Multimedia}, pages 3017--3025, 2022.

\bibitem[Nvidia(2021)]{NVENCVideoEncoder}
Nvidia.
\newblock {NVENC} {Video} {Encoder} {API} - {Emphasis} {MAP}, 2021.

\bibitem[Paigwar et~al.(2021)Paigwar, Sierra-Gonzalez, Erkent, and Laugier]{paigwar2021frustum}
Anshul Paigwar, David Sierra-Gonzalez, {\"O}zg{\"u}r Erkent, and Christian Laugier.
\newblock Frustum-pointpillars: A multi-stage approach for 3d object detection using rgb camera and lidar.
\newblock In \emph{Proceedings of the IEEE/CVF international conference on computer vision}, pages 2926--2933, 2021.

\bibitem[Qi et~al.(2018)Qi, Liu, Wu, Su, and Guibas]{qi2018frustum}
Charles~R Qi, Wei Liu, Chenxia Wu, Hao Su, and Leonidas~J Guibas.
\newblock Frustum pointnets for 3d object detection from rgb-d data.
\newblock In \emph{Proceedings of the IEEE conference on computer vision and pattern recognition}, pages 918--927, 2018.

\bibitem[Reich et~al.(2024)Reich, Debnath, Patel, Prangemeier, Cremers, and Chakradhar]{reich2024deep}
Christoph Reich, Biplob Debnath, Deep Patel, Tim Prangemeier, Daniel Cremers, and Srimat Chakradhar.
\newblock Deep video codec control for vision models.
\newblock In \emph{Proceedings of the IEEE/CVF Conference on Computer Vision and Pattern Recognition}, pages 5732--5741, 2024.

\bibitem[Richardson(2003)]{Richardson2003H2P}
Iain E.~Garden Richardson.
\newblock H. 264/mpeg-4 part 10 white peper : Transform \& quantization.
\newblock 2003.

\bibitem[Rudolph et~al.(2023)Rudolph, Schneegass, and Rizk]{rudolph2023rabbit}
Michael Rudolph, Stefan Schneegass, and Amr Rizk.
\newblock Rabbit: Live transcoding of v-pcc point cloud streams.
\newblock In \emph{Proceedings of the 14th Conference on ACM Multimedia Systems}, pages 97--107, 2023.

\bibitem[Shaheen et~al.(2022)Shaheen, Hanif, Hasan, and Shafique]{shaheen2022continual}
Khadija Shaheen, Muhammad~Abdullah Hanif, Osman Hasan, and Muhammad Shafique.
\newblock Continual learning for real-world autonomous systems: Algorithms, challenges and frameworks.
\newblock \emph{Journal of Intelligent \& Robotic Systems}, 105\penalty0 (1):\penalty0 9, 2022.

\bibitem[Shen and Gao(2021)]{shen2021rate}
Fangyu Shen and Wei Gao.
\newblock A rate control algorithm for video-based point cloud compression.
\newblock In \emph{2021 International Conference on Visual Communications and Image Processing (VCIP)}, pages 1--5. IEEE, 2021.

\bibitem[Sun et~al.(2020)Sun, Wang, Wang, Cheng, and Liu]{sun2020advanced}
Xuebin Sun, Sukai Wang, Miaohui Wang, Shing~Shin Cheng, and Ming Liu.
\newblock An advanced lidar point cloud sequence coding scheme for autonomous driving.
\newblock In \emph{Proceedings of the 28th ACM International Conference on Multimedia}, pages 2793--2801, 2020.

\bibitem[Tian et~al.(2017)Tian, Ochimizu, Feng, Cohen, and Vetro]{tianGeometricDistortionMetrics2017a}
Dong Tian, Hideaki Ochimizu, Chen Feng, Robert Cohen, and Anthony Vetro.
\newblock Geometric distortion metrics for point cloud compression.
\newblock In \emph{2017 {IEEE} {International} {Conference} on {Image} {Processing} ({ICIP})}, pages 3460--3464, 2017.
\newblock ISSN: 2381-8549.

\bibitem[Wang et~al.(2023)Wang, Zhu, Li, Xiao, and Liu]{wang2023vqba}
Shuoqian Wang, Mufeng Zhu, Na Li, Mengbai Xiao, and Yao Liu.
\newblock Vqba: Visual-quality-driven bit allocation for low-latency point cloud streaming.
\newblock In \emph{Proceedings of the 31st ACM International Conference on Multimedia}, pages 9143--9151, 2023.

\bibitem[Wang et~al.(2022)Wang, Wang, Liu, Jin, Jiang, and Chen]{wang2022enabling}
Yiding Wang, Weiyan Wang, Duowen Liu, Xin Jin, Junchen Jiang, and Kai Chen.
\newblock Enabling edge-cloud video analytics for robotics applications.
\newblock \emph{IEEE Transactions on Cloud Computing}, 11\penalty0 (2):\penalty0 1500--1513, 2022.

\bibitem[Wiegand et~al.(2003)Wiegand, Sullivan, Bjontegaard, and Luthra]{wiegand2003overview}
Thomas Wiegand, Gary~J Sullivan, Gisle Bjontegaard, and Ajay Luthra.
\newblock Overview of the h. 264/avc video coding standard.
\newblock \emph{IEEE Transactions on circuits and systems for video technology}, 13\penalty0 (7):\penalty0 560--576, 2003.

\bibitem[Wisultschew et~al.(2021)Wisultschew, Mujica, Lanza-Gutierrez, and Portilla]{wisultschew3DLIDARBasedObject2021}
Cristian Wisultschew, Gabriel Mujica, Jose~Manuel Lanza-Gutierrez, and Jorge Portilla.
\newblock {3D}-{LIDAR} {Based} {Object} {Detection} and {Tracking} on the {Edge} of {IoT} for {Railway} {Level} {Crossing}.
\newblock \emph{IEEE Access}, 9:\penalty0 35718--35729, 2021.
\newblock Conference Name: IEEE Access.

\bibitem[Xiao et~al.(2022)Xiao, Zhang, Wang, He, and Zhang]{xiao2022dnn}
Xuedou Xiao, Juecheng Zhang, Wei Wang, Jianhua He, and Qian Zhang.
\newblock Dnn-driven compressive offloading for edge-assisted semantic video segmentation.
\newblock In \emph{IEEE INFOCOM 2022-IEEE Conference on Computer Communications}, pages 1888--1897. IEEE, 2022.

\bibitem[Xie and Kim(2019)]{xie2019source}
Xiufeng Xie and Kyu-Han Kim.
\newblock Source compression with bounded dnn perception loss for iot edge computer vision.
\newblock In \emph{The 25th Annual International Conference on Mobile Computing and Networking}, pages 1--16, 2019.

\bibitem[Yang(2023)]{yang2023online}
Rui Yang.
\newblock \emph{Online continual learning for 3D detection of road participants in autonomous driving}.
\newblock PhD thesis, Universit{\'e} Bourgogne Franche-Comt{\'e}, 2023.

\bibitem[Yin et~al.(2021)Yin, Zhou, and Krahenbuhl]{yin2021center}
Tianwei Yin, Xingyi Zhou, and Philipp Krahenbuhl.
\newblock Center-based 3d object detection and tracking.
\newblock In \emph{Proceedings of the IEEE/CVF conference on computer vision and pattern recognition}, pages 11784--11793, 2021.

\bibitem[Zeng et~al.(2018)Zeng, Hu, Liu, Ye, Han, Li, and Sun]{zengRT3DRealTime3D2018}
Yiming Zeng, Yu Hu, Shice Liu, Jing Ye, Yinhe Han, Xiaowei Li, and Ninghui Sun.
\newblock {RT3D}: {Real}-{Time} 3-{D} {Vehicle} {Detection} in {LiDAR} {Point} {Cloud} for {Autonomous} {Driving}.
\newblock \emph{IEEE Robotics and Automation Letters}, 3\penalty0 (4):\penalty0 3434--3440, 2018.
\newblock Conference Name: IEEE Robotics and Automation Letters.

\bibitem[Zhang et~al.(2022)Zhang, Wang, and Liu]{zhang2022casva}
Miao Zhang, Fangxin Wang, and Jiangchuan Liu.
\newblock Casva: Configuration-adaptive streaming for live video analytics.
\newblock In \emph{IEEE INFOCOM 2022-IEEE Conference on Computer Communications}, pages 2168--2177. IEEE, 2022.

\bibitem[Zhao et~al.(2023)Zhao, Ning, Hong, Qiu, Lu, Zhao, Zhang, Zhou, Dai, Yang, et~al.]{zhao2023ada3d}
Tianchen Zhao, Xuefei Ning, Ke Hong, Zhongyuan Qiu, Pu Lu, Yali Zhao, Linfeng Zhang, Lipu Zhou, Guohao Dai, Huazhong Yang, et~al.
\newblock Ada3d: Exploiting the spatial redundancy with adaptive inference for efficient 3d object detection.
\newblock In \emph{Proceedings of the IEEE/CVF International Conference on Computer Vision}, pages 17728--17738, 2023.

\bibitem[Zhou et~al.(2019)Zhou, Wang, and Kr{\"a}henb{\"u}hl]{zhou2019objects}
Xingyi Zhou, Dequan Wang, and Philipp Kr{\"a}henb{\"u}hl.
\newblock Objects as points.
\newblock \emph{arXiv preprint arXiv:1904.07850}, 2019.

\end{thebibliography}
}

\clearpage
\setcounter{page}{1}
\maketitlesupplementary

\section{Details of RoI detectors}
\label{sup-roi-detector}
\subsection{Header Design}
Inspired by CenterPoint \cite{yin2021center}, we separate the heatmap predictions of objects with different sizes into 6 tasks. Backbone features first pass through a shared convolution block, which includes a $3 \times 3$ convolution layer, a batch normalization layer, and a ReLU activation layer. The output of the first convolution block is then fed into separate task-specific branches for each task. Each task branch contains a deformable convolution layer with a group size of 4, a convolution block with a $3 \times 3$ convolution layer, a batch normalization layer, and a ReLU layer, followed by a $3 \times 3$ convolution output layer.

\subsection{Training}
We implement our RoI detector in PyTorch using the open-sourced MMDetection3D~\cite{mmdet3d2020}. We train the model for 20 epochs for the RoI detector using the Adam~\cite{kingma2014adam} optimizer with a learning rate of $10^{-4}$, batch size of 9, and a weight decay of 0.01.

\section{Details of Back-end 3D Object Detectors}
\label{sup-backend-detector}
\subsection{Training}
We pre-train the back-end detectors with lossless point clouds following the settings in MMDetection3D~\cite{mmdet3d2020}. Specifically, we train each detector for 20 epochs using the Adam optimizer. The batch size is set to 9, the learning rate is set to $10^{-4}$, and the weight decay is set to 0.01. We then fine-tune detectors on a mixed dataset for one epoch using the same training parameters in the pre-training phase. The mixed dataset is a uniform mixture of training sets encoded by vanilla VPCC with QP $\in \{20, 25, 30\}$ and lossless encoding.
\subsection{Complexity}

Table.~\ref{tab-backend-detector-complexity} shows the complexities of the back-end detectors.

\begin{table}[ht]
    \centering
    \renewcommand{\arraystretch}{\TABVSPACE} 
    \setlength{\tabcolsep}{2pt} 
    \caption{Complexities of back-end detectors.}
    \begin{tabular}{c|c|c|c|c}
    \hline
     Model      & \multicolumn{2}{c|}{CenterPoint} & \multicolumn{2}{c}{BEVFusion-Lidar} \\
     
    \hline
      & FLOPs (G) & Param (M) &  FLOPs (G) & Param (M)  \\ 
\cline{2-3}         \cline{4-5}
    \hline
        Total & 119.33 & 6.11 & 85.98 & 5.15 \\ 
    \hline
    \end{tabular}
    \label{tab-backend-detector-complexity}
\end{table}

\section{Metric-bitrate Curves in Main Results}

\begin{figure}[t]
  \centering
    \includegraphics[width=\linewidth]{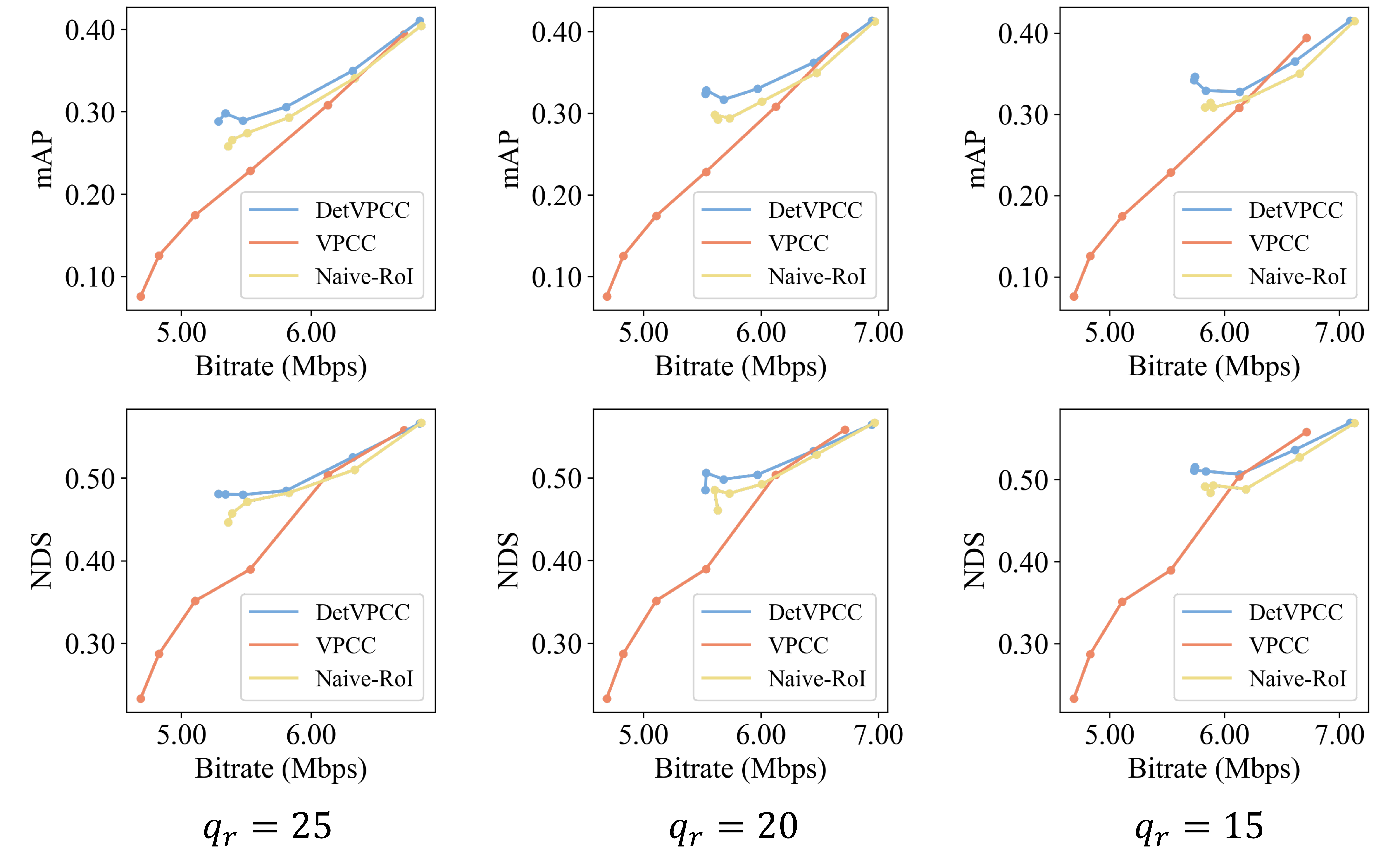}
  \caption[]{Metric-bitrate curve in \cref{sec-main-results} (CenterPoint)}
\label{fig-suppl-curve-centerpoint}
\end{figure}

\begin{figure}[t]
  \centering
    \includegraphics[width=\linewidth]{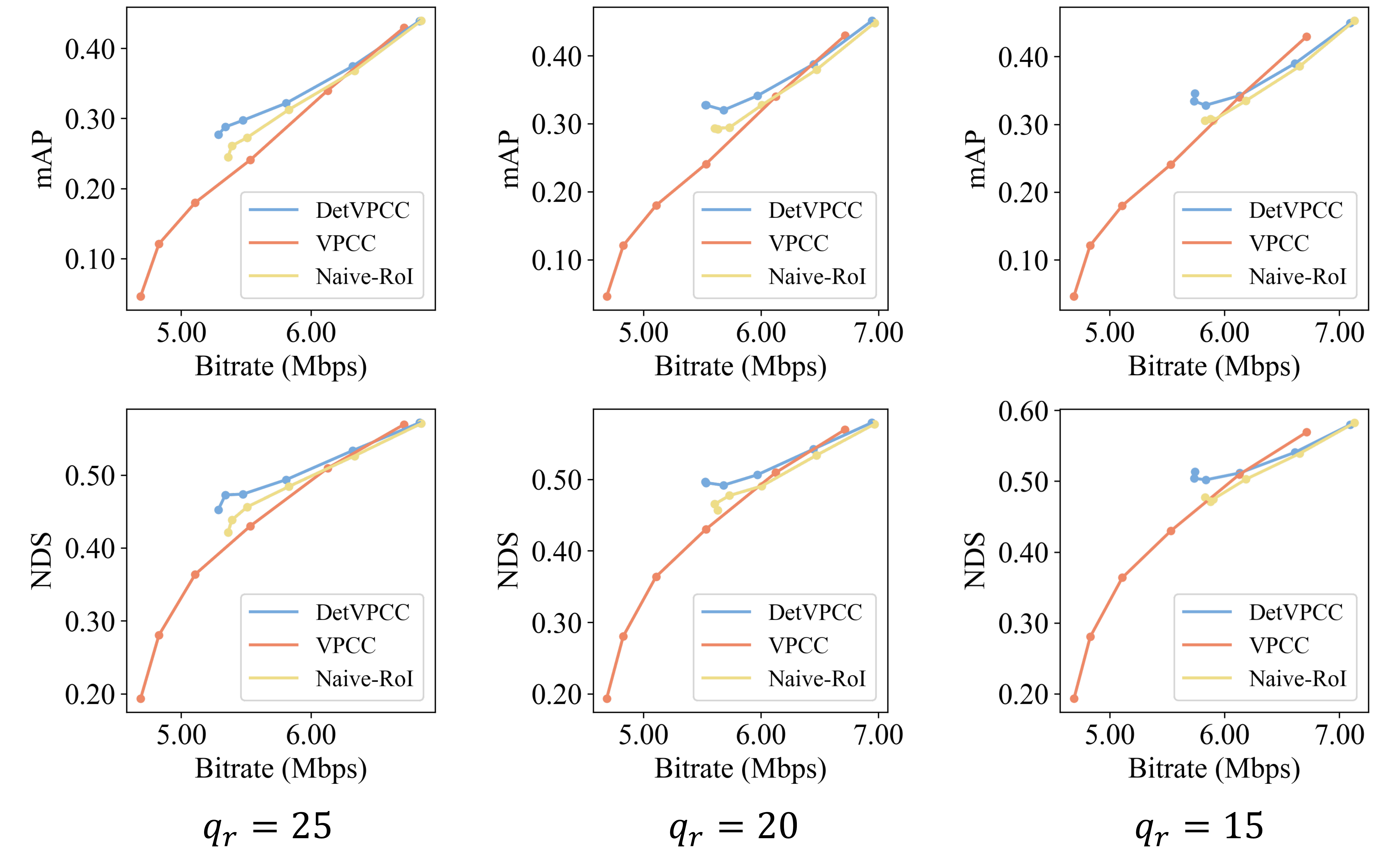}
  \caption[]{Metric-bitrate curve in \cref{sec-main-results} (BEVFusion-Lidar)}
\label{fig-suppl-curve-bevfusion}
\end{figure}

Fig.~\ref{fig-suppl-curve-centerpoint} and Fig.~\ref{fig-suppl-curve-bevfusion} give the mAP-bitrate and NDS-bitrate curves of \methodname{}, VPCC and Naive RoI in \cref{sec-main-results}.


\end{document}